\documentclass[conference]{IEEEtran}
\IEEEoverridecommandlockouts
\usepackage{cite}
\usepackage{amsmath,amssymb,amsfonts}
\usepackage{algorithmic}
\usepackage{multirow}
\usepackage{graphicx}
\usepackage{pdfpages}
\usepackage{fvextra}
\DefineVerbatimEnvironment{wrapverbatim}{Verbatim}{
  breaklines=true,
  breakanywhere=true,
  fontsize=\small
}
\usepackage{svg}
\usepackage{amsmath}
\usepackage{textcomp}
\usepackage{xcolor}
\usepackage{url}
\def\BibTeX{{\rm B\kern-.05em{\sc i\kern-.025em b}\kern-.08em
    T\kern-.1667em\lower.7ex\hbox{E}\kern-.125emX}}

\begin{document}

\title{Faithfulness metric fusion: Improving the evaluation of LLM trustworthiness across domains\\

\thanks{This work has been funded by the European Union. Views and opinions
expressed are however those of the authors only and do not necessarily
reflect those of the European Union or European Commission-EU. Neither
the European Union nor the granting authority can be held responsible for
them.}
}

\author{\IEEEauthorblockN{1\textsuperscript{st} Ben Malin}
\IEEEauthorblockA{\textit{CEDPS} \\
\textit{Brunel University London}\\
London, UK\\
Ben.Malin@brunel.ac.uk}
\and
\IEEEauthorblockN{2\textsuperscript{nd} Tatiana Kalganova}
\IEEEauthorblockA{\textit{CEDPS} \\
\textit{Brunel University London}\\
London, UK\\
Tatiana.Kalganova@brunel.ac.uk}
\and
\IEEEauthorblockN{3\textsuperscript{rd} Nikolaos Boulgouris}
\IEEEauthorblockA{\textit{CEDPS} \\
\textit{Brunel University London}\\
London, UK\\
Nikolaos.Boulgouris@brunel.ac.uk}

}

\maketitle

\begin{abstract}
We present a methodology for improving the accuracy of faithfulness evaluation in Large Language Models (LLMs). The proposed methodology is based on the combination of elementary faithfulness metrics into a combined (fused) metric, for the purpose of improving the faithfulness of LLM outputs. The proposed strategy for metric fusion deploys a tree-based model to identify the importance of each metric, which is driven by the integration of human judgements evaluating the faithfulness of LLM responses. This fused metric is demonstrated to correlate more strongly with human judgements across all tested domains for faithfulness. Improving the ability to evaluate the faithfulness of LLMs, allows for greater confidence to be placed within models, allowing for their implementation in a greater diversity of scenarios. Additionally, we homogenise a collection of datasets across question answering and dialogue-based domains and implement human judgements and LLM responses within this dataset, allowing for the reproduction and trialling of faithfulness evaluation across domains.
\end{abstract}


\section{Introduction}
Faithfulness within the realm of Large Language Models (LLMs) refers to their ability to produce trustworthy and reliable outputs \cite{Malin-faithfulness-journal-paper}. This has major implications in the domains that they can be applied within, as it has been noted that LLMs have a tendency to "hallucinate" outputs and provide nonfactual information to the user \cite{li-etal-2024-dawn}. These "hallucinations" diminish the applicability of LLMs in many contexts, most notably when related to safety-critical applications whereby erroneous information can incur major consequences. Evaluating the faithfulness of LLM outputs is a growing field due to this desirability for LLM reliability and user-confidence. There are a wide-range of metrics that have been commonly utilised in the assessment of faithfulness, with some being more prevalent within specific domains (such as summarisation or question-answering).

The ability to reliably evaluate faithfulness can allow for greater confidence to be placed in LLMs, whilst also allowing for alternative approaches to be taken in directly improving faithfulness. Most notably this can refer to reprompting strategies, whereby the LLM is tasked with regenerating the output given additional context, with the aim of mitigating the risk of hallucination.
This paper focuses on evaluating LLM faithfulness using a question-answering dataset across a range of commonplace metrics, as well as applying faithfulness-improving approaches that have been demonstrated by other authors, in a novel domain. This is conducted with the intention of building confidence in LLM outputs, so that they can be implemented in a wider-range of scenarios whereby responsibility is key and potentially unethical outcomes can be avoided.
The novelties introduced within this paper are as follows: 
\begin{itemize}
\item The implementation of graph-based metrics across question-answering domains. 
\item Fusion of faithfulness metrics to improve correlation with human judgement.
\item Production of a cross-domain dataset featuring human evaluations regarding faithfulness.
\end{itemize}
\section{Related work}

Within the faithfulness evaluation field, there are several commonplace metrics that have been used across a variety of domains.
Human judgements regarding faithfulness are commonly gathered so that they can be used to validate novel metrics and approaches, typically through correlation \cite{wang2023evaluatingopenqaevaluation,adlakha2024evaluatingcorrectnessfaithfulnessinstructionfollowing}. This methodological approach is considered the gold standard for faithfulness evaluation. However, this can be a time-consuming and lengthy process, and is not feasible for many applications due to these constraints.

N-gram metrics \cite{lin-2004-rouge} are ubiquitous within the literature, this is largely due to the low computational requirements, as many studies have showcased weaker correlations with human judgement than has been seen with the other metrics listed here \cite{falke-etal-2019-ranking,Goodrich2019AssessingText,Kryscinski2019EvaluatingSummarization}.
This is largely because a small number of n-grams convey the majority of the information, and synonymous n-grams are considered to be different despite being semantically similar \cite{wang-etal-2020-asking}.

BERTScore \cite{Zhang2019BERTScore} is another common metric within this field and has been shown to outperform n-grams in many studies. BERTScore (and other similar metrics) leverage pretrained embeddings to identify the proximity between source and reference text, and is able to capture semantic information due to the composition of the embeddings.

Exact match (EM) identifies whether the output is identical to the ground truth, whilst Lexical Match (LM) requires the output to contain the ground truth. Both metrics have featured heavily within the evaluation of faithfulness for question-answering (QA) tasks \cite{wang2023evaluatingopenqaevaluation,adlakha2024evaluatingcorrectnessfaithfulnessinstructionfollowing}. However, EM has been noted as being too strict and misclassifying faithful outputs. This is alleviated somewhat with LM \cite{yao2024accuratenuancedopenqaevaluation}, but both strategies typically require several ground truth answers so that aliases and synonyms can be correctly classified. However, both metrics have demonstrated high correlations with human judgment in a number of studies\cite{wang2023evaluatingopenqaevaluation,adlakha2024evaluatingcorrectnessfaithfulnessinstructionfollowing}.
Joint Goal Accuracy (JGA) is  a commonly used metric for the evaluation of dialogue state tracking \cite{dey-etal-2022-towards}. It functions comparably to EM, evaluating predicted slot values against ground truth slot values within a dialogue environment. Comparably to EM, it is also a strict faithfulness measure, with several critiques that relate to inflexibility for accepting synonymous values as well as misclassifications from earlier turns impeding the results in subsequent dialogue turns \cite{dey-etal-2022-towards,yang-etal-2022-multi}.

LLMs have been used to evaluate the faithfulness of generative outputs within question-answering and summarization domains, with varied implementations. This is commonly referred to as LLM-as-a-judge \cite{chen-etal-2024-humans}. Some authors have incorporated ground truth data within the LLM prompt, which can aid in its judgment, and other authors retrieve granular outputs (such as 5-point or 100-point scales) instead of a binary classification \cite{fabbri2021summevalreevaluatingsummarizationevaluation, fu-etal-2023-large,Kocmi2021ToTranslation}. Both the inclusion of ground truth data as well as increased output granularity have been shown to improve evaluation capabilities. However, several studies have demonstrated an over-confidence of LLMs in classifying textual data as faithful, which is a key element of study within this paper \cite{wang2023evaluatingopenqaevaluation,adlakha2024evaluatingcorrectnessfaithfulnessinstructionfollowing}. Additionally, the LLM evaluation metric is the only one of these that can function without the provision of ground-truth data, either using a Retrieval Augmented Generation (RAG) approach or through using innate knowledge, though this can affect reliability \cite{alinejad2024}. This is most notable within machine translation, whereby Quality Estimation (QE) \cite{Blatz2004ConfidenceTranslation}is a common approach for assessing faithfulness across languages \cite{Bererd2023HowTranslation}.

Distinct from these metrics is graph generation, a technique which extracts entities and relationships from unstructured text to produce a graph, which can then be evaluated using more traditional metrics. This has been shown to bolster faithfulness evaluation within the summarisation domain, evidenced through improvements in correlation with human judgment \cite{kim2024fablesevaluatingfaithfulnesscontent,ribeiro-etal-2022-factgraph}.
This technique has had varied implementations but functions in comparable ways, with various graph generation strategies having been trialled, with Abstract Meaning Representation graphs most commonly being used\cite{ribeiro-etal-2022-factgraph}.

This data structuring approach is also aligned with other works which have extracted relationships from unstructured text for evaluation, even if graphs are not directly produced \cite{Goodrich2019AssessingText}.
Through the extraction of key entities and relationships, alternative metrics can then be used, with the structured data intending to lessen the difficulty of evaluation. SMatch \cite{cai-knight-2013-smatch} is one such metric, functioning comparably to EM/LM metrics, identifying matches between two graphs to calculate an overall degree of similarity using recall and precision.

Despite other authors having demonstrated the successes that can be attained using these data structuring approaches for evaluation, to the best of our knowledge, they have only been trialed within the summarisation domain, despite being relevant elsewhere.

\section{Datasets}
The scope of this study covers faithfulness evaluation for dialogue summarisation, as well as question-answering domains, with human judgements for LLM responses being provided within a fused dataset for these domains. 
Both domains are desirable to include in this study due to the prevalence of faithfulness evaluation within these domains, allowing for novel methodologies to be more appropriately compared \cite{wang2023evaluatingopenqaevaluation,adlakha2024evaluatingcorrectnessfaithfulnessinstructionfollowing, fabbri2021summevalreevaluatingsummarizationevaluation,wang-etal-2020-asking}. 
Additionally, Task-Oriented Dialogue (TOD) datasets are incorporated within the fused dataset, due to  the data being aligned with conversational evaluation. However, TOD datasets are less suitable for "faithfulness" evaluation, as datasets within this domain are typically evaluated using strict outputs, with no ambiguity. Although, commonly used metrics within this domain, such as JGA, have raised queries regarding their limitations \cite{dey-etal-2022-towards,feng-etal-2023-towards,yang-etal-2022-multi}. Due to these queries, and to allow for more rigorous evaluation across a broader TOD dataset, this domain is implemented within the fused dataset.

We fuse and homogenize datasets across these domains, whilst incorporating human factuality judgments and LLM responses\footnote{\url{https://huggingface.co/datasets/Brunel-AI/ELOQUENCE}}. The dataset has been compartmentalised with the faithfulness task being used to separate the various datasets - allowing for greater homogeneity of schema, where schema in this context refers to the attributes and data types used within the datasets.

\subsection{Task Oriented Dialogues}
Many TOD datasets have unique schema requirements, with alternative data included as well as differing representations. This can be highlighted through slots and slot values, which exist in all identified TOD datasets, but can be represented with varied levels of nesting or data types\cite{zang-etal-2020-multiwoz,rastogi2020scalablemultidomainconversationalagents}. A slot represents a category of key information in completing the task, and the slot value represents the specific value attached to this category. Furthermore, attributes such as intent or external knowledge are not universally available. Due to these discrepancies among reviewed datasets, the homogenization of schema will allow for a more varied and larger evaluation dataset. The datasets selected for fusion are SGD \cite{rastogi2020scalablemultidomainconversationalagents}, DSTC3 \cite{DSTC3Henderson2014}, SpokenWOZ \cite{spokenwoz2023} and MultiWOZ2.2\cite{zang-etal-2020-multiwoz}. These datasets have been chosen due to them conveying unique attributes such as audio or multilinguality.
\begin{figure*}[]
  \centering
  \includegraphics[width=0.6\textwidth]{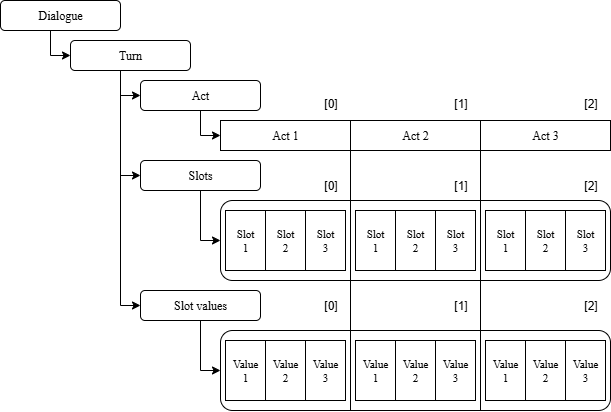}
  \caption{Task-Oriented Dialogue Nesting Illustration}
  \label{fig:tod-nesting}
  
\end{figure*}
\subsection{Question Answering}
 Natural Questions (NQ) \cite{kwiatkowski-etal-2019-natural} has been selected due to the presence of long-form and short-form answers, providing greater scope for the assessment of the suitability of fact extraction techniques. In addition, the PopQA dataset \cite{mallen-etal-2023-trust} is incorporated into the fused dataset due to the presence of fact tuples. However, only entity-based answers are present, limiting the evaluation of long-form faithfulness for this dataset. Similarly, the QAConv \cite{wu-etal-2022-qaconv} dataset has been incorporated within the fused dataset for extractive QA evaluation, and features multi-turn dialogue between several speakers with QA pairs relating to the dialogue. However, evaluation of this subset of QA is outside of the scope of this study.

 \subsection{Dialogue Summarisation}
The dialogue summarisation domain involves long-form text, often in the form of multi-turn dialogue between multiple speakers, with the key points summarised in a concise manner. These summaries are typically human produced, and can feature the key points separated (which can be referred to as atomic facts \cite{min-etal-2023-factscore}) or more typically as a paragraph. The SAMSum dataset \cite{gliwa-etal-2019-samsum} has been selected for fusion into the dialogue summarisation domain of this fused dataset due to the presence of human annotated summaries and multi-turn dialogue. 

  \subsection{Conversational QA}
  The final domain for which faithfulness evaluation is experimentally evaluated is conversational QA, which contains dialogue between multiple speakers, with QA pairs relating to the dialogue. This is desirable for evaluation, as  LLMs cannot use inherent knowledge to answer questions, and thus it is more closely aligned with reading comprehension tasks. Additionally, the relevant text that is required for processing is significantly larger than in non-conversational QA domains, further validating the selected metrics across different use cases. To evaluate this domain, the QA Conv. dataset \cite{wu-etal-2022-qaconv} has been selected and utilised within the fused dataset. It contains multi-turn dialogues that are separated by speaker, as well as a question per dialogue, with one or more correct answers per question. Human judgements have been collected and appended within the fused dataset's conversational QA subset for evaluation.
  
\subsection{Dataset Processing}
The following methodology is conducted to process the datasets gathered, so that they can be homogenised with regards to schema, as well as to format the data to be more conducive to faithfulness evaluation.

Samples are removed from NQ \cite{kwiatkowski-etal-2019-natural}when multiple long-form ground-truth samples are available, simplifying the evaluation and mitigating the likelihood of exceeding token limits. Additionally, QA pairs are only retained when all annotators agree on the long-form answer (or only one is present), whilst all short-form answers are retained. All ground-truth answers containing HTML markup are removed (most commonly tables), with the sample removed if this process results in an absence of a ground-truth. Samples in PopQA \cite{mallen-etal-2023-trust} that contain invalid UTF-8 characters are removed for consistency. Furthermore, data within PopQA is homogenised with NQ through the merging of the ground-truth answers with their aliases. Answer spans and turn ids have been extracted and appended to the QAConv \cite{wu-etal-2022-qaconv} subset of the fused dataset, as well as the sources relevant to the QA pairs being appended to a single data point for more streamlines usage.

Primarily, the proposed schema for the TOD portion of the fused dataset is based upon MultiWOZ 2.2, a heavily validated and widely used dataset within this domain \cite{zang-etal-2020-multiwoz}. However, in addition to the attributes present within MultiWOZ 2.2, a summary dictionary is appended for each sample, containing the full dialogue with relevant slots, acts and utterances. The motivation behind this is the future assessment of dialogue summarization, as the provision of turn-level data and slot values can provide additional utility.
Due to the variation in schema across the TOD datasets, relationships between acts, intents, slots and slot values can all be represented in different ways. To homogenize these relationships for the fused dataset, slots and slot values are nested into lists which correlate to their relevant act type. This can lead to an unnecessary degree of nesting for specific datasets which can only feature one act type per slot, but it is necessary to ensure no data is lost and for consistency across the dataset. This nesting is explained through Figure \ref{fig:tod-nesting} which showcases a small portion of the attributes within the TOD subset, with each act having corresponding Slots and Slot values that occur at the same index position. Through this procedure, 4500 multi-turn dialogues have been aggregated within this dataset.

The SAMSum \cite{gliwa-etal-2019-samsum} dataset has been formatted to identify the key points within the summarised paragraph, through sentence extraction. This methodology allows for LLM-generated summarised points to be evaluated against each ground truth summarised point, for pairwise metrics - which have been shown to increase correlation with human judgement as well as increasing the granularity of the faithfulness metrics \cite{min-etal-2023-factscore}.

\subsection{Human Evaluation}
Within the NQ \cite{kwiatkowski-etal-2019-natural} subset, long-form and short-form LLM responses have been evaluated against their respective ground-truth answers, with human judgments provided which consider the context of the question and the alignment between response and answer. When there is uncertainty over the faithfulness of a response, the human assessors can abstain from providing a judgment and these samples are not taken into account for the correlation. It is important to note that the ground-truths may not always be factual (commonly due to outdated information within the dataset), but should be treated as such and thus external knowledge is not used to deem a response to be faithful if it is not aligned with the ground-truth.
Human judgements have been collected using a Python script which displays the question, any additional context (such as a dialogue transcript), the ground truth answers/summaries as well as the LLM-generated answers/summaries. For all domains a score of 1-5 is annotated for the LLM response, this is then additionally represented as a human boolean value, whereby scores of 3-5 are considered a 1 (faithful) and scores below 3 are scored as a 0 (unfaithful). For the summarisation domain, humans perform this judgement for each LLM summarised point, with the mean then taken and used to represent the overall faithfulness of the LLM's response. This process is utilised due to the additional ambiguity found within the summarisation domain, as well as to allow for evaluation of individual facts within a larger summary - which has demonstrated superior correlation with human judgement in several studies \cite{Kim2024FABLESEF,min-etal-2023-factscore}.
\subsection{Fused dataset}
The final dataset comprises four subsets, each of which have been processed in the aims of improving data quality, with the dataset statistics as follows:
The question answering domain contains 500 QA pairs, with LLM responses, generated graphs, automated metrics and 100 human judgements assigned. The conversational QA domain also covers 500 samples, with 500 multi-turn dialogues, associated QA pairs, with LLM responses, generated graphs and automated metrics, with 100 human judgements assigned. The dialogue summarisation section contains 500 multi-turn dialogues, with their human annotated summaries \cite{gliwa-etal-2019-samsum}, along with LLM summaries, generated graphs, automated metrics and 100 human judgements assigned per sample (with each sample containing several human judgements).
Task-Oriented Dialogues contains 4,500 homogenised samples across the discussed datasets, though no human judgements, metrics or LLM responses have been appended to this subset.

\section{Dataset generation methodology}
\subsection{LLM}
The finetuned Llama 3.1-8B instruct model has been used to generate responses for the various tasks within the dataset, utilising a greedy search strategy (ensuring that outputs are deterministic) and setting the maximum new tokens to 256. All tasks use an in-context learning approach, whereby example outputs are incorporated into the prompt, improving the consistency of the generated outputs.
The prompts used differ across the tasks, and all prompts are included within the appendix.

To increase the granularity of the faithfulness evaluation that the LLM provides, two approaches have been trialled. The first utilises logits to determine the confidence of the faithfulness prediction, and subsequently alter the score from a boolean measurement to a continuous measurement between 0 and 1. The second approach prompts the LLM to output the degree of faithfulness on a 5-point Likert scale. These are referred to as the LLM Confidence approach and the LLM Likert approach respectively Both of these techniques have been assessed before, and been shown to improve faithfulness evaluation capabilities \cite{Kim2024FABLESEF,subbiah-etal-2024-storysumm, fu-etal-2023-large,tsvilodub2024predictionslanguagemodelsmultiplechoice}. Furthermore, the speakers have been identified across the dialogue and formatted as a separate attribute.

\subsection{AMR Graph Generation}
\begin{figure}[]
  \centering
\includegraphics[width=0.5\textwidth]{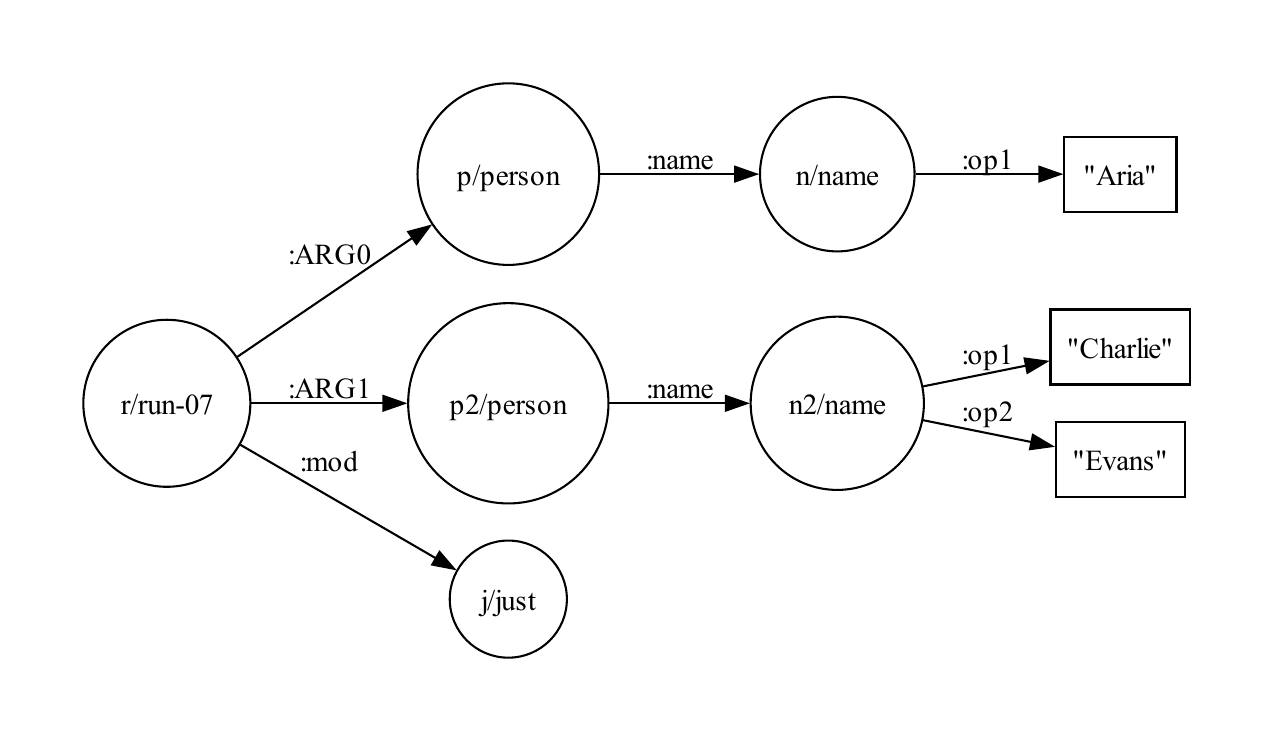}
  \caption{SAMSum AMR Sample}
  \label{fig:samsum-amr_sample}
  
\end{figure}

\begin{figure*}[]
  \centering
\includegraphics[width=0.8\textwidth]{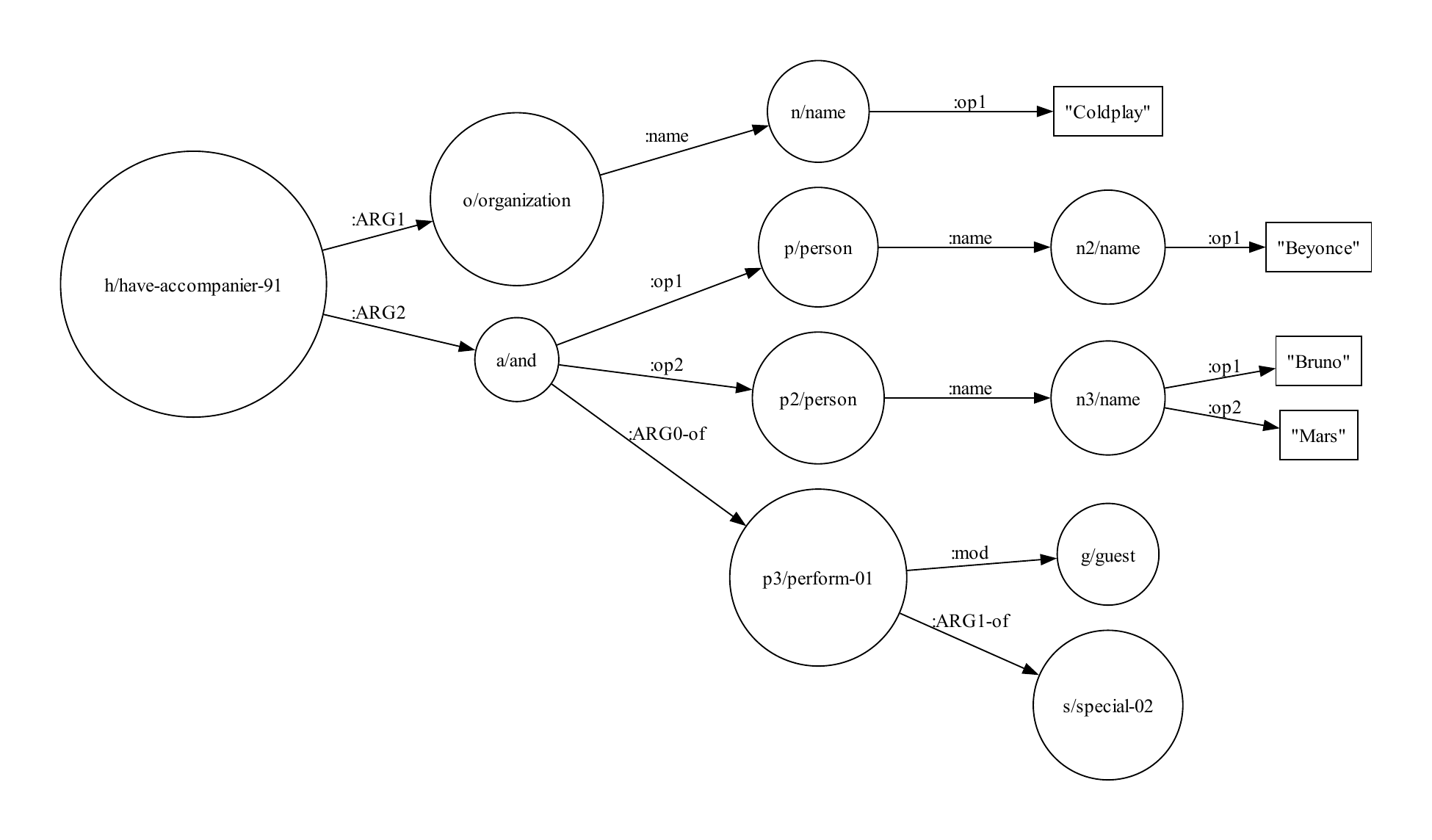}
  \caption{NaturalQuestions AMR Sample}
  \label{fig:NQ-amr_sample}
  
\end{figure*}

For the assessment of graph generation approaches we have used STOG BART Large \cite{zhang-etal-2018-stog}, a BART model trained on the AMR-3 dataset \cite{amr-3}. This technique is only trialled for the QA and dialogue summarisation domains, as TOD already utilises highly structured data and thus is less suitable for this approach. 
For QA, the reference graphs are generated using both long-form and short-form ground truth answers and are evaluated against the long-form and short-form LLM responses respectively. For dialogue summarisation, reference graphs are generated using both the full dialogue history, as well as the ground truth summaries. These graphs are then evaluated against the LLM-generated summary graphs.
The metrics used for the graph evaluation are the same for both domains, with SMatch being calculated across all evaluations. However, SMatch scores are calculated with additional configurations also - such as F1 score for named entities (Entity F1), the removal of Word-Sense-Disambiguation (No WSD F1) and the calculation of F1 without relationship labels being necessitated (Unlabeled F1). These additional metrics allow for greater scope of evaluation. When several graphs are produced, we take the mean, maximum and minimum score for each of these metrics to cover a wider range of possible identified features.
Figure \ref{fig:samsum-amr_sample} has been generated from the SAMSum subset of the fused dataset, and conveys the original summary text: 'Aria has just run into Charlie Evans.'.
Figure \ref{fig:NQ-amr_sample} has been generated from the Natural Questions subset using the ground truth short-form answer: 'Coldplay with special guest performers Beyonce and Bruno Mars'


\section{Results}
\subsection{Question Answering}
Common metrics for QA faithfulness evaluation are calculated using the short-form and long-form answers and responses, as well as for short-form and long-form fact-tuples. Two LLM metrics are trialled, the LLM Likert approach which prompts the model to output a faithfulness score between 0 and 5, with 0 representing complete unfaithfulness and 5 representing perfect alignment between the ground-truth and the response. The second LLM metric assessed is LLM Conf., which prompts the model to output a binary faithfulness score, which is then reduced by the token logits in the case of a faithful prediction or increased for an unfaithful prediction. This provides additional granularity for faithfulness through utilising the probability of the classification.

We find that the long-form QA task to favour the LLM confidence approach (denoted as LLM Conf. within the tables), whilst the short-form evaluation favours the Likert approach as demonstrated by the highest correlations achieved within Table \ref{tab: Question Answering Correlations}, though this is a minor difference for long-form. Across most metrics, short-form evaluation produces higher correlations than their long-form counterparts. This is to be expected, and is likely a large reason why QA faithfulness evaluation typically utilizes these short-form datasets. The highest correlation attained overall is found using the short-form domain, when using the LLM Likert approach, with a correlation of 0.815.
We observe that the graph-based metrics perform comparably to n-gram strategies for long-form evaluation, yet these are not strong correlations. Whilst the n-gram metrics attain strong correlations for short-form, unlike graph-based metrics. The Exact Match metric only features correlations for short-form QA due to no matches being made for long-form QA, resulting in no derived correlation.However, lexical match was the best non-LLM metric for long-form evaluation, though for short-form these correlations were surpassed by n-grams and BERTScore. The correlations attained for both long-form and short-form QA approaches are detailed below in Table \ref{tab: Question Answering Correlations}.
 \begin{table*}[h]
\centering
\caption{Question answering correlations with human judgement}
\label{tab: Question Answering Correlations}
\begin{tabular}{|c|ccccc|cc|cc|}
\hline
\multirow{2}{*}{\textit{\textbf{Metric}}} & \multicolumn{5}{c|}{\textit{\textbf{Metric Classification}}}                                                                                                                                                              & \multicolumn{2}{c|}{\textit{\textbf{Short-Form}}}                                                       & \multicolumn{2}{c|}{\textit{\textbf{Long-Form}}}                                                        \\ \cline{2-10} 
                                          & \multicolumn{1}{c|}{\textit{\textbf{LLM}}} & \multicolumn{1}{c|}{\textit{\textbf{N-gram}}} & \multicolumn{1}{c|}{\textit{\textbf{Graph}}} & \multicolumn{1}{c|}{\textit{\textbf{Embedding}}} & \textit{\textbf{Matching}} & \multicolumn{1}{l|}{\textit{\textbf{Human Likert}}} & \multicolumn{1}{l|}{\textit{\textbf{Human Bool}}} & \multicolumn{1}{l|}{\textit{\textbf{Human Likert}}} & \multicolumn{1}{l|}{\textit{\textbf{Human Bool}}} \\ \hline
\textit{LLM Likert}                       & \multicolumn{1}{c|}{\textit{X}}            & \multicolumn{1}{c|}{\textit{}}                & \multicolumn{1}{c|}{\textit{}}               & \multicolumn{1}{c|}{\textit{}}                   & \textit{}                  & \multicolumn{1}{c|}{\textbf{0.815}}                 & \textbf{0.783}                                    & \multicolumn{1}{c|}{0.658}                          & \textbf{0.642}                                    \\ \hline
\textit{LLM Conf.}                        & \multicolumn{1}{c|}{\textit{X}}            & \multicolumn{1}{c|}{\textit{}}                & \multicolumn{1}{c|}{\textit{}}               & \multicolumn{1}{c|}{\textit{}}                   & \textit{}                  & \multicolumn{1}{c|}{0.729}                          & 0.728                                             & \multicolumn{1}{c|}{\textbf{0.659}}                 & 0.637                                             \\ \hline
\textit{ROUGE-1}                          & \multicolumn{1}{c|}{\textit{}}             & \multicolumn{1}{c|}{\textit{X}}               & \multicolumn{1}{c|}{\textit{}}               & \multicolumn{1}{c|}{\textit{}}                   & \textit{}                  & \multicolumn{1}{c|}{0.629}                          & 0.617                                             & \multicolumn{1}{c|}{0.235}                          & 0.191                                             \\ \hline
\textit{ROUGE-L}                          & \multicolumn{1}{c|}{\textit{}}             & \multicolumn{1}{c|}{\textit{X}}               & \multicolumn{1}{c|}{\textit{}}               & \multicolumn{1}{c|}{\textit{}}                   & \textit{}                  & \multicolumn{1}{c|}{0.623}                          & 0.608                                             & \multicolumn{1}{c|}{0.240}                          & 0.189                                             \\ \hline
\textit{BERTScore}                        & \multicolumn{1}{c|}{\textit{}}             & \multicolumn{1}{c|}{\textit{}}                & \multicolumn{1}{c|}{\textit{}}               & \multicolumn{1}{c|}{\textit{X}}                  & \textit{}                  & \multicolumn{1}{c|}{0.502}                          & 0.500                                             & \multicolumn{1}{c|}{0.378}                          & 0.350                                             \\ \hline
\textit{Lexical match}                    & \multicolumn{1}{c|}{\textit{}}             & \multicolumn{1}{c|}{\textit{}}                & \multicolumn{1}{c|}{\textit{}}               & \multicolumn{1}{c|}{\textit{}}                   & \textit{X}                 & \multicolumn{1}{c|}{0.457}                          & 0.427                                             & \multicolumn{1}{c|}{0.388}                          & 0.357                                             \\ \hline
\textit{ROUGE-2}                          & \multicolumn{1}{c|}{\textit{}}             & \multicolumn{1}{c|}{\textit{X}}               & \multicolumn{1}{c|}{\textit{}}               & \multicolumn{1}{c|}{\textit{}}                   & \textit{}                  & \multicolumn{1}{c|}{0.451}                          & 0.434                                             & \multicolumn{1}{c|}{0.277}                          & 0.228                                             \\ \hline
\textit{Unlabeled F1 mean}                & \multicolumn{1}{c|}{\textit{}}             & \multicolumn{1}{c|}{\textit{}}                & \multicolumn{1}{c|}{\textit{X}}              & \multicolumn{1}{c|}{\textit{}}                   & \textit{}                  & \multicolumn{1}{c|}{0.203}                          & 0.223                                             & \multicolumn{1}{c|}{0.245}                          & 0.228                                             \\ \hline
\textit{SMATCH mean}                      & \multicolumn{1}{c|}{\textit{}}             & \multicolumn{1}{c|}{\textit{}}                & \multicolumn{1}{c|}{\textit{X}}              & \multicolumn{1}{c|}{\textit{}}                   & \textit{}                  & \multicolumn{1}{c|}{0.183}                          & 0.191                                             & \multicolumn{1}{c|}{0.228}                          & 0.195                                             \\ \hline
\textit{No WSD F1   mean}                 & \multicolumn{1}{c|}{\textit{}}             & \multicolumn{1}{c|}{\textit{}}                & \multicolumn{1}{c|}{\textit{X}}              & \multicolumn{1}{c|}{\textit{}}                   & \textit{}                  & \multicolumn{1}{c|}{0.182}                          & 0.191                                             & \multicolumn{1}{c|}{0.231}                          & 0.200                                             \\ \hline
\textit{Unlabeled F1 min}                 & \multicolumn{1}{c|}{\textit{}}             & \multicolumn{1}{c|}{\textit{}}                & \multicolumn{1}{c|}{\textit{X}}              & \multicolumn{1}{c|}{\textit{}}                   & \textit{}                  & \multicolumn{1}{c|}{0.178}                          & 0.195                                             & \multicolumn{1}{c|}{0.224}                          & 0.210                                             \\ \hline
\textit{SMATCH max}                       & \multicolumn{1}{c|}{\textit{}}             & \multicolumn{1}{c|}{\textit{}}                & \multicolumn{1}{c|}{\textit{X}}              & \multicolumn{1}{c|}{\textit{}}                   & \textit{}                  & \multicolumn{1}{c|}{0.176}                          & 0.185                                             & \multicolumn{1}{c|}{0.238}                          & 0.200                                             \\ \hline
\textit{No WSD F1 max}                    & \multicolumn{1}{c|}{\textit{}}             & \multicolumn{1}{c|}{\textit{}}                & \multicolumn{1}{c|}{\textit{X}}              & \multicolumn{1}{c|}{\textit{}}                   & \textit{}                  & \multicolumn{1}{c|}{0.174}                          & 0.183                                             & \multicolumn{1}{c|}{0.244}                          & 0.210                                             \\ \hline
\textit{Unlabeled F1 max}                 & \multicolumn{1}{c|}{\textit{}}             & \multicolumn{1}{c|}{\textit{}}                & \multicolumn{1}{c|}{\textit{X}}              & \multicolumn{1}{c|}{\textit{}}                   & \textit{}                  & \multicolumn{1}{c|}{0.172}                          & 0.192                                             & \multicolumn{1}{c|}{0.257}                          & 0.238                                             \\ \hline
\textit{Exact match}                      & \multicolumn{1}{c|}{\textit{}}             & \multicolumn{1}{c|}{\textit{}}                & \multicolumn{1}{c|}{\textit{}}               & \multicolumn{1}{c|}{\textit{}}                   & \textit{X}                 & \multicolumn{1}{c|}{0.161}                          & 0.171                                             & \multicolumn{1}{c|}{N/A}                            & N/A                                               \\ \hline
\textit{SMATCH min}                       & \multicolumn{1}{c|}{\textit{}}             & \multicolumn{1}{c|}{\textit{}}                & \multicolumn{1}{c|}{\textit{X}}              & \multicolumn{1}{c|}{\textit{}}                   & \textit{}                  & \multicolumn{1}{c|}{0.154}                          & 0.160                                             & \multicolumn{1}{c|}{0.214}                          & 0.186                                             \\ \hline
\textit{No WSD F1 min}                    & \multicolumn{1}{c|}{\textit{}}             & \multicolumn{1}{c|}{\textit{}}                & \multicolumn{1}{c|}{\textit{X}}              & \multicolumn{1}{c|}{\textit{}}                   & \textit{}                  & \multicolumn{1}{c|}{0.153}                          & 0.159                                             & \multicolumn{1}{c|}{0.213}                          & 0.187                                             \\ \hline
\textit{Entity F1 min}                    & \multicolumn{1}{c|}{\textit{}}             & \multicolumn{1}{c|}{\textit{}}                & \multicolumn{1}{c|}{\textit{X}}              & \multicolumn{1}{c|}{\textit{}}                   & \textit{}                  & \multicolumn{1}{c|}{-0.057}                         & -0.045                                            & \multicolumn{1}{c|}{-0.111}                         & -0.161                                            \\ \hline
\textit{Entity F1   mean}                 & \multicolumn{1}{c|}{\textit{}}             & \multicolumn{1}{c|}{\textit{}}                & \multicolumn{1}{c|}{\textit{X}}              & \multicolumn{1}{c|}{\textit{}}                   & \textit{}                  & \multicolumn{1}{c|}{-0.086}                         & -0.078                                            & \multicolumn{1}{c|}{-0.092}                         & -0.141                                            \\ \hline
\textit{Entity F1 max}                    & \multicolumn{1}{c|}{\textit{}}             & \multicolumn{1}{c|}{\textit{}}                & \multicolumn{1}{c|}{\textit{X}}              & \multicolumn{1}{c|}{\textit{}}                   & \textit{}                  & \multicolumn{1}{c|}{-0.127}                         & -0.127                                            & \multicolumn{1}{c|}{-0.066}                         & -0.112                                            \\ \hline
\end{tabular}
\end{table*}
 \vspace{2mm}

\subsection{Conversational QA}
The conversational QA domain functions similarly to the QA domain outlined previously. The primary difference is the incorporation of context within the prompt, taking the form of inputting the entire dialogue history prior to the question. This provision of context within the prompt makes this task comparable to a reading comprehension task. These LLM responses are then appended to the fused dataset for the metrics to be calculated and incorporated, including LLM-as-a-judge approaches. The human judgements that are gathered for this domain are produced in the same manner as with the previous QA task.
We find lexical match to be the most highly correlated metric with human judgement when evaluating for a Likert score, achieving a correlation of 0.461. Although, no metrics perform as well as with the previous domain. It is worth noting the significant reduction in correlation when using the LLM Likert approach compared to the LLM Conf. approach. Some explanation to the lower correlations found with this domain can be evidenced by ambiguous and strangely worded questions, such as "Which time of day does NPR White House correspondent Tamara Keith Ellison and federal workers finally getting back to work have in common ?" with the answer "For the first time". It is questions such as this where the ground truth answers can be misleading and potentially decrease correlations identified.

  The correlations attained for the conversational QA domain are detailed below in Table \ref{tab:conversationalqa correlations}.

  \begin{table*}[h]
\centering
\caption{Conversational QA correlations with human judgement}
\label{tab:conversationalqa correlations}
\begin{tabular}{|c|ccccc|l|l|}
\hline
\multirow{2}{*}{\textit{\textbf{Metric}}} & \multicolumn{5}{c|}{\textit{\textbf{Metric Classification}}}                                                                                                                                                              & \multicolumn{1}{c|}{\multirow{2}{*}{\textit{\textbf{Human Likert}}}} & \multicolumn{1}{c|}{\multirow{2}{*}{\textit{\textbf{Human Bool}}}} \\ \cline{2-6}
                                          & \multicolumn{1}{c|}{\textit{\textbf{LLM}}} & \multicolumn{1}{c|}{\textit{\textbf{N-gram}}} & \multicolumn{1}{c|}{\textit{\textbf{Graph}}} & \multicolumn{1}{c|}{\textit{\textbf{Embedding}}} & \textit{\textbf{Matching}} & \multicolumn{1}{c|}{}                                                & \multicolumn{1}{c|}{}                                              \\ \hline
\textit{Lexical match}                    & \multicolumn{1}{l|}{}                      & \multicolumn{1}{l|}{}                         & \multicolumn{1}{l|}{}                        & \multicolumn{1}{l|}{}                            & \multicolumn{1}{l|}{X}     & \textbf{0.461}                                                       & 0.429                                                              \\ \hline
\textit{LLM Conf.}                        & \multicolumn{1}{c|}{\textit{X}}            & \multicolumn{1}{c|}{\textit{}}                & \multicolumn{1}{c|}{\textit{}}               & \multicolumn{1}{c|}{\textit{}}                   & \textit{}                  & 0.427                                                                & \textbf{0.453}                                                     \\ \hline
\textit{ROUGE-1}                          & \multicolumn{1}{c|}{\textit{X}}            & \multicolumn{1}{c|}{\textit{}}                & \multicolumn{1}{c|}{\textit{}}               & \multicolumn{1}{c|}{\textit{}}                   & \textit{}                  & 0.403                                                                & 0.377                                                              \\ \hline
\textit{ROUGE-L}                          & \multicolumn{1}{c|}{\textit{}}             & \multicolumn{1}{c|}{\textit{X}}               & \multicolumn{1}{c|}{\textit{}}               & \multicolumn{1}{c|}{\textit{}}                   & \textit{}                  & 0.399                                                                & 0.374                                                              \\ \hline
\textit{ROUGE-2}                          & \multicolumn{1}{c|}{\textit{}}             & \multicolumn{1}{c|}{\textit{X}}               & \multicolumn{1}{c|}{\textit{}}               & \multicolumn{1}{c|}{\textit{}}                   & \textit{}                  & 0.285                                                                & 0.270                                                              \\ \hline
\textit{BERTScore}                        & \multicolumn{1}{c|}{\textit{}}             & \multicolumn{1}{c|}{\textit{}}                & \multicolumn{1}{c|}{\textit{}}               & \multicolumn{1}{c|}{\textit{X}}                  & \textit{}                  & 0.272                                                                & 0.268                                                              \\ \hline
\textit{LLM Likert}                       & \multicolumn{1}{c|}{\textit{}}             & \multicolumn{1}{c|}{\textit{}}                & \multicolumn{1}{c|}{\textit{}}               & \multicolumn{1}{c|}{\textit{}}                   & \textit{X}                 & 0.232                                                                & 0.213                                                              \\ \hline
\textit{Unlabeled F1}                     & \multicolumn{1}{c|}{\textit{}}             & \multicolumn{1}{c|}{\textit{X}}               & \multicolumn{1}{c|}{\textit{}}               & \multicolumn{1}{c|}{\textit{}}                   & \textit{}                  & 0.199                                                                & 0.188                                                              \\ \hline
\textit{SMATCH}                           & \multicolumn{1}{c|}{\textit{}}             & \multicolumn{1}{c|}{\textit{}}                & \multicolumn{1}{c|}{\textit{X}}              & \multicolumn{1}{c|}{\textit{}}                   & \textit{}                  & 0.192                                                                & 0.182                                                              \\ \hline
\textit{No WSD F1}                        & \multicolumn{1}{c|}{\textit{}}             & \multicolumn{1}{c|}{\textit{}}                & \multicolumn{1}{c|}{\textit{X}}              & \multicolumn{1}{c|}{\textit{}}                   & \textit{}                  & 0.191                                                                & 0.181                                                              \\ \hline
\textit{Entity F1}                        & \multicolumn{1}{c|}{\textit{}}             & \multicolumn{1}{c|}{\textit{}}                & \multicolumn{1}{c|}{\textit{X}}              & \multicolumn{1}{c|}{\textit{}}                   & \textit{}                  & 0.068                                                                & 0.053                                                              \\ \hline
\textit{Exact match}                      & \multicolumn{1}{c|}{\textit{}}             & \multicolumn{1}{c|}{\textit{}}                & \multicolumn{1}{c|}{\textit{X}}              & \multicolumn{1}{c|}{\textit{}}                   & \textit{}                  & 0.052                                                                & 0.046                                                              \\ \hline
\end{tabular}
\end{table*}
 \vspace{2mm}
\subsection{Dialogue Summarisation}
Prompting the LLM to produce summarised points from the multi-turn dialogue and evaluating these responses against the ground truth summary as well as the full dialogue transcript is conducted for the evaluation of faithfulness. 
The SAMSum \cite{gliwa-etal-2019-samsum} dataset has been selected for use in this faithfulness evaluation. The LLM-as-a-judge techniques are aligned with the previously used methodologies, relating to the Likert prompting approach and the use of token confidence to influence the faithfulness score.
For evaluating against the ground truth, this is conducted in a pairwise manner, whereby the generated summarised points are evaluated against the extracted ground-truth summarised points, with the highest scoring match being determined to be the paired summary point. This provides a score for each summary point made by the LLM, which are then used to calculated mean, maximum and minimum scores for each metric. This is considered the "fact"-based approach. This technique was developed through inspiration from Min et al., with their FACTScore approach, which has been demonstrated in the literature to improve correlation with human judgement compared to long-form comparison \cite{min-etal-2023-factscore}.
This contrasts the "Full" approach, whereby the LLM judgement is produced to cover all summarised points. Additional metrics were trialed for the graph-based approaches, with graphs being generated for each dialogue turn (to be evaluated in a pairwise manner with both the transcript, as well as the ground truth summaries), as well as graphs being evaluated using the full transcript graph.
Additionally, metrics have been calculated using the full dialogue transcript for the LLM and graph-based metrics - due to their capability in assessing long-form text.

The correlations attained for both ground truth evaluation and transcript strategies are detailed below in Table \ref{tab:Summarisation Correlations}. Due to the additional range of metrics computed for this domain, only positively correlated metrics are included within the table.
We find the best-performing metric to be the LLM Conf. approach that utilises a faithfulness scoring for each generated summary (the "fact"-based approach) and then averaged. This is aligned with the manner in which the human judgements have been recorded for correlating. Interestingly, the second best metric is the same approach yet with the overall score being determined by the least faithful point. This is likely due to the generaly prevalence for LLMs to overestimate faithfulness, which can attribute too much importance to an averaged score. The use of this "fact"-based approach significantly outperforms the use of LLM's to score all of the summarised points as a whole. However, this could be due to the fact that the human evaluations have been gathered in a similar way, with correlations being calculated against the mean human judgement for all LLM summaries. It is of note that when the correlations are calculated using the human boolean mean (with Likert scores of three or more being considered faithful, and the average taken for all judgements) that the correlations are significantly weaker, with very few significant correlations being identified. The metrics that are not LLM-based suffer a significant drop in performance. Additionally, we opt to use the mean score of human judgement for calculating correlations, as opposed to maximums or minimums, due to the desire for determining faithfulness across all summaries, as opposed to a singular point.

\begin{table*}[h]
\centering
\caption{Summarisation correlations with human judgement}
\label{tab:Summarisation Correlations}
\begin{tabular}{|c|ccccc|l|l|}
\hline
\multirow{2}{*}{\textbf{Metric}}      & \multicolumn{5}{c|}{\textit{\textbf{Metric Classification}}}                                                                                                                                                              & \multicolumn{1}{c|}{\multirow{2}{*}{\textit{\textbf{Human Likert mean}}}} & \multicolumn{1}{c|}{\multirow{2}{*}{\textit{\textbf{Human Bool mean}}}} \\ \cline{2-6}
                                      & \multicolumn{1}{c|}{\textit{\textbf{LLM}}} & \multicolumn{1}{c|}{\textit{\textbf{N-gram}}} & \multicolumn{1}{c|}{\textit{\textbf{Graph}}} & \multicolumn{1}{c|}{\textit{\textbf{Embedding}}} & \textit{\textbf{Matching}} & \multicolumn{1}{c|}{}                                                     & \multicolumn{1}{c|}{}                                                   \\ \hline
\textit{LLM Conf. Fact mean}          & \multicolumn{1}{l|}{X}                     & \multicolumn{1}{l|}{}                         & \multicolumn{1}{l|}{}                        & \multicolumn{1}{l|}{}                            & \multicolumn{1}{l|}{}      & \textbf{0.578}                                                            & 0.380                                                                   \\ \hline
\textit{LLM Conf. Fact min}           & \multicolumn{1}{c|}{\textit{X}}            & \multicolumn{1}{c|}{\textit{}}                & \multicolumn{1}{c|}{\textit{}}               & \multicolumn{1}{c|}{\textit{}}                   & \textit{}                  & 0.573                                                                     & \textbf{0.415}                                                          \\ \hline
\textit{LLM Likert Fact mean}         & \multicolumn{1}{c|}{\textit{X}}            & \multicolumn{1}{c|}{\textit{}}                & \multicolumn{1}{c|}{\textit{}}               & \multicolumn{1}{c|}{\textit{}}                   & \textit{}                  & 0.465                                                                     & 0.264                                                                   \\ \hline
\textit{LLM Likert Fact min}          & \multicolumn{1}{c|}{\textit{X}}            & \multicolumn{1}{c|}{\textit{}}                & \multicolumn{1}{c|}{\textit{}}               & \multicolumn{1}{c|}{\textit{}}                   & \textit{}                  & 0.409                                                                     & 0.322                                                                   \\ \hline
\textit{BERTScore mean}               & \multicolumn{1}{c|}{\textit{}}             & \multicolumn{1}{c|}{\textit{}}                & \multicolumn{1}{c|}{\textit{}}               & \multicolumn{1}{c|}{\textit{X}}                  & \textit{}                  & 0.221                                                                     & 0.135                                                                   \\ \hline
\textit{BERTScore max}                & \multicolumn{1}{c|}{\textit{}}             & \multicolumn{1}{c|}{\textit{}}                & \multicolumn{1}{c|}{\textit{}}               & \multicolumn{1}{c|}{\textit{X}}                  & \textit{}                  & 0.199                                                                     & 0.143                                                                   \\ \hline
\textit{LLM Likert}                   & \multicolumn{1}{c|}{\textit{X}}            & \multicolumn{1}{c|}{\textit{}}                & \multicolumn{1}{c|}{\textit{}}               & \multicolumn{1}{c|}{\textit{}}                   & \textit{}                  & 0.164                                                                     & 0.023                                                                   \\ \hline
\textit{Transcript Unlabeled F1 max}  & \multicolumn{1}{c|}{\textit{}}             & \multicolumn{1}{c|}{\textit{}}                & \multicolumn{1}{c|}{\textit{X}}              & \multicolumn{1}{c|}{\textit{}}                   & \textit{}                  & 0.164                                                                     & 0.149                                                                   \\ \hline
\textit{Transcript Unlabeled F1 min}  & \multicolumn{1}{c|}{\textit{}}             & \multicolumn{1}{c|}{\textit{}}                & \multicolumn{1}{c|}{\textit{X}}              & \multicolumn{1}{c|}{\textit{}}                   & \textit{}                  & 0.164                                                                     & 0.149                                                                   \\ \hline
\textit{Transcript Unlabeled F1 mean} & \multicolumn{1}{c|}{\textit{}}             & \multicolumn{1}{c|}{\textit{}}                & \multicolumn{1}{c|}{\textit{X}}              & \multicolumn{1}{c|}{\textit{}}                   & \textit{}                  & 0.164                                                                     & 0.149                                                                   \\ \hline
\textit{Transcript SMATCH max}        & \multicolumn{1}{c|}{\textit{}}             & \multicolumn{1}{c|}{\textit{}}                & \multicolumn{1}{c|}{\textit{X}}              & \multicolumn{1}{c|}{\textit{}}                   & \textit{}                  & 0.150                                                                     & 0.149                                                                   \\ \hline
\textit{Transcript SMATCH min}        & \multicolumn{1}{c|}{\textit{}}             & \multicolumn{1}{c|}{\textit{}}                & \multicolumn{1}{c|}{\textit{X}}              & \multicolumn{1}{c|}{\textit{}}                   & \textit{}                  & 0.150                                                                     & 0.149                                                                   \\ \hline
\textit{Transcript SMATCH mean}       & \multicolumn{1}{l|}{}                      & \multicolumn{1}{l|}{}                         & \multicolumn{1}{l|}{X}                       & \multicolumn{1}{c|}{\textit{}}                   & \textit{}                  & 0.150                                                                     & 0.149                                                                   \\ \hline
\textit{ROUGE-2 mean}                 & \multicolumn{1}{l|}{}                      & \multicolumn{1}{l|}{X}                        & \multicolumn{1}{l|}{}                        & \multicolumn{1}{c|}{\textit{}}                   & \textit{}                  & 0.148                                                                     & 0.090                                                                   \\ \hline
\textit{ROUGE-2 max}                  & \multicolumn{1}{l|}{}                      & \multicolumn{1}{l|}{X}                        & \multicolumn{1}{l|}{}                        & \multicolumn{1}{c|}{\textit{}}                   & \textit{}                  & 0.140                                                                     & 0.059                                                                   \\ \hline
\textit{Turn GT SMATCH max}           & \multicolumn{1}{l|}{}                      & \multicolumn{1}{l|}{}                         & \multicolumn{1}{l|}{X}                       & \multicolumn{1}{l|}{}                            & \multicolumn{1}{l|}{}      & 0.136                                                                     & 0.159                                                                   \\ \hline
\textit{BERTScore min}                & \multicolumn{1}{l|}{}                      & \multicolumn{1}{l|}{}                         & \multicolumn{1}{l|}{}                        & \multicolumn{1}{l|}{X}                           & \multicolumn{1}{l|}{}      & 0.136                                                                     & 0.055                                                                   \\ \hline
\textit{ROUGE-L max}                  & \multicolumn{1}{l|}{}                      & \multicolumn{1}{l|}{X}                        & \multicolumn{1}{l|}{}                        & \multicolumn{1}{l|}{}                            & \multicolumn{1}{l|}{}      & 0.128                                                                     & 0.088                                                                   \\ \hline
\textit{Turn GT Unlabeled F1 mean}    & \multicolumn{1}{l|}{}                      & \multicolumn{1}{l|}{}                         & \multicolumn{1}{l|}{X}                       & \multicolumn{1}{l|}{}                            & \multicolumn{1}{l|}{}      & 0.127                                                                     & 0.148                                                                   \\ \hline
\textit{Turn GT Unlabeled F1 max}     & \multicolumn{1}{l|}{}                      & \multicolumn{1}{l|}{}                         & \multicolumn{1}{l|}{X}                       & \multicolumn{1}{l|}{}                            & \multicolumn{1}{l|}{}      & 0.123                                                                     & 0.149                                                                   \\ \hline
\end{tabular}
\end{table*}
 \vspace{2mm}

\section{Metric Fusion}
In the aims of improving correlation with human judgement we aimed to incorporate several metrics together, due to the unique ways in which many of them function. Each metric is able to capture different features between the ground truth and generated texts, based on how they operate - such as n-gram approaches focusing on the lexical structure of the text, whilst LLMs offer a more abstractive metric with higher-level reasoning capabilities. The range of metrics that are evaluated can thus identify a wider-range of features and be less prone to issues that can affect singular metrics. Similarly, it has been demonstrated that LLM-as-a-judge metrics favour to judge responses as faithful, whereas matching metrics incur the opposite bias, with the fusion of these metrics we aim to improve upon these drawbacks that are specific to individual metrics \cite{Malin-faithfulness-journal-paper}. To fuse these metrics we use the Explainable Boosting Machine (EBM) \cite{interpretml} algorithm, which is used to weight the metrics in the aims of maximising similarity with the human judgement. This algorithm operates in a more explainable way than other algorithms such as Random Forest (RF) \cite{random-forest-2001}, with the feature importances calculated able to be used directly with the scores outputted to compute the output variable - which in our case is a faithfulness metric. This weighted metric is then tested on a blind subset of the relevant domains.

\subsection{Fused metric weights}
The weightings of the fused metric has been provided in Table \ref{tab:fused metric weights}, detailing the importance of each metric in the optimisation of human correlation, as identified through the EBM network. Considering the weak correlations displayed by graph-based metrics, they are consistently deemed highly important for the fused metric, with the exception of short-form QA. Reasons for this are likely due to the comparative simplicity within short-form evaluation, with graphs likely being superfluous for evaluation. However, within the summarisation domain (which contains the highest level of ambiguity and the longest form text) it is shown that graph-based metrics are exceedingly important, which is unsurprising considering that it is the domain that pioneered this approach. Furthermore, this importance is likely increased due to the matching metrics not being relevant, with no matches (neither exact nor lexical) being made, as well as BERTScore being identified as not important to the aligning with faithfulness.

\begin{table*}[]
\centering
\caption{EBM derived fused metric weightings}
\label{tab:fused metric weights}
\begin{tabular}{|l|c|c|c|c|}
\hline
\textit{\textbf{Metric}} & \multicolumn{1}{l|}{\textit{\textbf{Short-Form QA}}} & \multicolumn{1}{l|}{\textit{\textbf{Long-Form QA}}} & \multicolumn{1}{l|}{\textit{\textbf{Conversational   QA}}} & \multicolumn{1}{l|}{\textit{\textbf{Summarisation}}} \\ \hline
\textit{LLM}             & 0.171                                                & 0.194                                               & 0.148                                                      & 0.263                                                \\ \hline
\textit{N-gram}          & 0.298                                                & 0.172                                               & \textbf{0.295}                                             & 0.114                                                \\ \hline
\textit{Embedding}       & 0.094                                                & 0.075                                               & 0.078                                                      & 0.134                                                \\ \hline
\textit{Graph}           & \textbf{0.307}                                       & \textbf{0.370}                                      & 0.244                                                      & \textbf{0.434}                                       \\ \hline
\textit{Matching}        & 0.131                                                & 0.189                                               & 0.234                                                      & N/A                                                  \\ \hline
\end{tabular}
\end{table*}
\subsection{Fused metric evaluation}

Table \ref{tab:fused metric correlations} details and compares the correlations with human judgement that each metric attains, as well as the fused metric which incorporates an assortment of the metrics in aim of improving correlation. For these correlations, we have opted to correlate for the human likert judgement, for the sake of brevity. Additionally, it is worth noting that the correlations achieved for the fused metric is gathered using a blind test set, to ensure that the test is fair. In all cases, the correlation with human judgement is improved when utilising a fused metric. However, the gains that can be attained are varied depending on the domain, with long-form QA boasting an improvement of 0.136, yet it's short-form counterpart only improved by 0.005.

Through application of the EBM algorithm for identifying the level of importance that each metric has, we can observe that LLM-based metrics and n-gram based metrics provide the most utility for a fused metric. For long-form QA, the graph based metrics contribute to 37.0\% of the fused metric's outputted value, with the LLM-based techniques adding an additional 19.4\% combined. Whilst for short-form QA, n-gram metrics contributes 29.8\% and the LLM-based approaches contribute an additional 17.1\%. The significantly higher importance assigned to the n-gram metrics for short-form QA is likely due to the greater concision with which responses and answers are provided. The conversational QA fused metric recognised the n-gram metrics to be the most important for improving the correlation with human judgement, weighting 29.5\% of the whole. Finally, for the summarisation domain, the graph-based metrics comprised the highest weighting, with 43.4\%. For this domain, matching metrics never attained a match, explaining their exclusion from the weights and the correlations.

\begin{table*}[]
\centering
\caption{Evaluation of the proposed fused faithfulness metrics}
\label{tab:fused metric correlations}
\begin{tabular}{|l|l|l|l|l|}
\hline
\textit{\textbf{Metric}} & \textit{\textbf{Short-Form QA}} & \textit{\textbf{Long-Form QA}} & \textit{\textbf{Conversational   QA}} & \textit{\textbf{Summarisation}} \\ \hline
\textit{Prev. Best}      & \multicolumn{1}{c|}{0.815}      & \multicolumn{1}{c|}{0.659}     & 0.461                                 & 0.578                           \\ \hline
\textit{Fused Metric}    & \textbf{0.820}                  & \textbf{0.795}                 & \textbf{0.507}                        & \textbf{0.581}                  \\ \hline
\end{tabular}
\end{table*}

\section{Discussion}
LLM-as-a-judge metrics displayed the greatest consistency across domains, with the LLM Conf. approach which uses token logits to provide a continuous faithfulness value, being in the top-2 performing metrics across all domains.
The difficulty of evaluating summarisations for faithfulness is further evidenced within this study, with only LLM-based techniques attaining strong correlations, and graph-based techniques attaining mild-moderate correlations. All other metrics exhibit weak correlations with humans, demonstrating their incompatibility for such a domain. This is likely due to summaries containing similar entities, yet presenting different semantic meaning.
Graph-based performance is underwhelming for all domains with regards to it's individual correlation with human judgement. However, its utility is displayed when used as part of a larger combinatorial metric, evidenced by the weighting shown in \ref{tab:fused metric weights}. 
These results have also demonstrated the increased reliability of evaluating short-form text compared to long-form text, with all metrics that attain a significant correlation having superior performance when transitioning to short-form QA from long-form QA. Additionally, our results show that there is utility in the application of a fused metric to improve the correlation with human judgement, though the improvements that can be expected are domain-dependent.

\section{Conclusion}
The level of adaptability across domains offered by LLM-as-a-judge techniques makes them incredibly versatile. Additionally, they commonly attain the strongest correlation with human judgements, proving that more trust can be placed within these metrics compared to some of the more traditional metrics, even domain-specific metrics. Increasing the level of granularity of LLM faithfulness outputs improved the correlation with human judgement, with two methodologies tested: prompting the LLMs to output Likert scores of 1-5, and requesting binary faithfulness scores which are then converted to continuous values through token logits.
We also find that for QA tasks, the vast majority of metrics are more faithful evaluators when answers are concise, with very strong correlations observed for LLM-as-a-judge techniques for short-form QA. However, we do find that upon fusion of several metrics into a single faithfulness metric through use of the Explainable Boosting Machine that the more ambiguous, long-form domain is able to attain comparable performance - with a correlation of 0.795.
The use of a fused faithfulness metric has demonstrated the ability to improve on correlation with human judgement across all domains, with differing weightings that are tailored to each.
Graph-based metrics, whilst not highly correlating on their own have been shown to be very important within the fused metric in all domains, though when answers are longer and more ambiguous (such as with Long-form QA and dialogue summarisation) this importance is especially strong.
It is identified that the summarisation and conversational QA domains are the most challenging, with a wide range of metrics and approaches being evaluated, yet the vast majority do not attain significant correlations. However, through the fusion of some of these metrics we attain our best-performing metric for both domains, improving correlation from 0.578 to 0.581 and 0.461 to 0.507 respectively. These domains are more challenging due to being dialogue-oriented, with greater ambiguity regarding the faithfulness of their answers.

\bibliographystyle{IEEEtran}
\bibliography{references-2.bib}
\begin{appendices}
\section{Question Answering Prompts}
\subsection{LLM Answering}
\begin{wrapverbatim}
    
Long form prompt = (
'You will be given a trivia question, your task is to answer this question to the best of your knowledge.'
'Please produce enough text to suitably answer the question whilst covering all key information, but only answer the question. Aim to succinctly answer within 2 sentences.'
"Do NOT provide any text additional to the generated answer.")
Short form prompt = (
'You will be given a trivia question, your task is to answer this question to the best of your knowledge.'
'Please produce your answer concisely, aiming to give only the key information necessary.'
"Do NOT provide any text additional to the generated answer.")
\end{wrapverbatim}

\subsection{LLM Evaluation}
\begin{wrapverbatim}
LLM Likert prompt = (
"You will be given a trivia question, as well as the ground truth answer and a 
user's response.""Your task is to provide
a rating on how well the user's answer 
aligns with the ground truth answer, 
given the context of the question."
"The ground truth answer is always correct,
do not use external knowledge to dispute 
this.""Give your answer as an integer on 
a scale of 0 to 5, where 0 means that the
user's answer is completely incorrect, a
score of 3 should be used when their 
answer is somewhat correct but may be 
missing additional information."
"A score of 5 means that the user's answer
on it's own is correct and answers the 
question.""Do NOT provide any text 
additional to the score."
""
"Here is the question, ground truth answer 
and user's answer:"
""

f"Question: {question}"
f"Ground truth: {ground_truth_answer}"
f"User answer: {response}")


LLM Confidence  = (
"You will be given a trivia question, as well as the ground truth answer and a user's response."
"Your task is to provide a rating on how well the user's answer aligns with the ground truth answer, given the context of the question."
"The ground truth answer is always correct, do not use external knowledge to dispute this."
"Give your answer as either Factual or Not-Factual. Not-Factual means that the user's answer is incorrect."
"A rating of Factual means that the user's answer is correct."
"Do NOT provide any text additional to the rating."
""
"Here is the question, ground truth answer and user's answer:"
""
f"Question: {question}"
f"Ground truth: {ground_truth_answer}"
f"User answer: {response}")
       \end{wrapverbatim}

\section{Conversational Question Answering Prompts}
\subsection{LLM Answering}
\begin{wrapverbatim}

'You will be given a dialogue between two or more people as well as a question that pertains to the dialogue, your task is to answer the question to the best of your ability.'
'Please produce a concise answer that succinctly answers the question. Do not provide information that is out of scope of the asked question.'

\end{wrapverbatim}

\subsection{LLM Evaluation}
\begin{wrapverbatim}

Evaluation system prompt = ("You are a Large Language Model with the primary objective of providing factually"
  " correct ratings regarding the alignment of two answers, given the provided context.")

Likert evaluation prompt = ("You will be given a transcript of a dialogue between two or more people, as well as a question, the ground truth answer and a user's answer."
"Your task is to provide a rating on how well the user's answer aligns with the ground truth answer, given the context of the dialogue transcript."
"Give your answer as an integer on a scale of 0 to 5, where 0 means that the user's answer convey significantly erroneous information, a score of 3 should be used when some key information is missing or there are minor factual inconsistencies or amibugity within the answer."
"A score of 5 means that all important information is conveyed with no ambiguity or factual inconsistency."
"Do NOT provide any text additional to the score."
""
"Here is the transcript with the question, ground truth answer and user's answer:"
""
f"Transcript: {transcript}"
f"Ground truth answer: {gt_answer}"
f"User answer: {user_answer}")
Confidence evaluation prompt = ("You will be given a transcript of a dialogue between two or more people, as well as a question, the ground truth answer and a user's answer."
"Your task is to provide a rating on how well the user's answer aligns with the ground truth answer, given the context of the dialogue transcript."
"Give your answer as a value of either Faithful or Not-Faithful, where Not-Faithful means that the user's answer convey significantly erroneous information."
"A score of Faithful means that all important information is conveyed with no ambiguity or factual inconsistency."
"Do NOT provide any text additional to the rating."
""
"Here is the transcript with the question, ground truth answer and user's answer:"
""
f"Transcript: {transcript}"
f"Ground truth answer: {gt_answer}"
f"User answer: {user_answer}")
\end{wrapverbatim}

\section{Dialogue Summarisation Prompts} 
\subsection{LLM Answering}
\begin{wrapverbatim}

'You will be given a dialogue between two or more people, your task is to summarise this dialogue, through the production of summary points.'
'Please produce a few sentences, summarising the individual key points that were discussed in the provided interaction.'
"Do NOT provide any text additional to the summary. Use a '*' to mark each summarised point, and this should be the first character in the generated response.")
\end{wrapverbatim}

\subsection{LLM Evaluation}
\begin{wrapverbatim}

System prompt = ("You are a Large Language Model with the primary objective of providing factually"
" correct ratings regarding the alignment of summarised dialogues.")
LLM Likert evaluation fact based prompt = ("You will be given a transcript of a dialogue between two or more people, and a single key point within this dialogue."
"Your task is to provide a rating on how well the summarised point aligns with the full dialogue."
"Give your answer as an integer on a scale of 0 to 5, where 0 means that the summarised point is not present within the full dialogue or is significantly misleading, a score of 3 should be used when the point contains a minor inconsistency or level of ambiguity."
"A score of 5 means that the summarised point is completely correct and unambiguous with respect to the dialogue."
"Do NOT provide any text additional to the score."
""
"Here is the transcript and summarised points:"
""
f"Transcript: {transcript}"
f"Summarised point: {summary_point}")
LLM Confidence evaluation fact based promp = ("You will be given a transcript of a dialogue between two or more people, and a single key point within this dialogue."
"Your task is to provide a rating on how well the summarised point aligns with the full dialogue."
"Give your answer as a value of either Faithful or Not-Faithful, where Not-Faithful means that the summarised point is not present within the full dialogue or is significantly misleading"
"A rating of Faithful means that the summarised point is completely correct and unambiguous with respect to the dialogue."
"Do NOT provide any text additional to the rating."
""
"Here is the transcript and summarised points:"
""
f"Transcript: {transcript}"
f"Summarised point: {summary_point}")
LLM Likert Evaluation "Full" = ("You will be given a transcript of a dialogue between two or more people, and a summary of the key points within this dialogue."
"Your task is to provide a rating on how well the summarised points align with the full dialogue."
"Give your answer as an integer on a scale of 0 to 5, where 0 means that the summarised points convey significantly erroneous information, a score of 3 should be used when some key information is missing or there are minor factual inconsistencies or amibugity within the summary."
"A score of 5 means that all important information is conveyed with no ambiguity or factual inconsistency."
"Do NOT provide any text additional to the score."
""
"Here is the transcript and summarised points:"
""
f"Transcript: {transcript}"
f"Summarised points: {summary}")
LLM Confidence Evaluation "Full" = ("You will be given a transcript of a dialogue between two or more people, and a summary of the key points within this dialogue."
"Your task is to provide a rating on how well the summarised points align with the full dialogue."
"Give your answer as a value of either Faithful or Not-Faithful, where Not-Faithful means that the summarised points convey significantly erroneous information."
"A rating of Faithful means that all important information is conveyed with no ambiguity or factual inconsistency."
"Do NOT provide any text additional to the rating."
""
"Here is the transcript and summarised points:"
""
f"Transcript: {transcript}"
f"Summarised points: {summary}")
\end{wrapverbatim}
\end{appendices}

\vspace{12pt}

\end{document}